\documentclass[oneside]{article}

\usepackage[final]{neurips_2024}

\usepackage[utf8]{inputenc} 
\usepackage[T1]{fontenc}    
\usepackage{hyperref}       
\usepackage{url}            
\usepackage{booktabs}       
\usepackage{amsfonts}       
\usepackage{nicefrac}       
\usepackage{microtype}      
\usepackage{xcolor}         
\usepackage{threeparttable}
\usepackage{ulem}
\usepackage{amsmath}
\usepackage{algorithm}
\usepackage{algorithmic}
\usepackage{mdframed}
\usepackage{graphicx} 
\usepackage{amssymb}
\usepackage{multirow}
\usepackage{tabularx}
\usepackage{booktabs}
\usepackage{geometry}
\usepackage{longtable}
\usepackage{times,a4wide}
\usepackage{graphicx}
\usepackage{amsmath}
\usepackage{listings}
\usepackage{tabularx}
\usepackage[most]{tcolorbox}

\usepackage{listings}
\usepackage{float}

\title{A Full-duplex Speech Dialogue Scheme Based On Large Language Model}

\newcommand*\samethanks[1][\value{footnote}]{\footnotemark[#1]}
\author{
	Peng Wang\thanks{Equal contribution.}\quad  
	Songshuo Lu\samethanks{} \quad
	Yaohua Tang\samethanks{}
	 \\
	\textbf{
		Sijie Yan \quad
        Wei Xia   \quad
		Yuanjun Xiong} 
	\\\\
	MThreads AI
	\\
	\texttt{w8ngp1ng@gmail.com, lusongshuo97@gmail.com, tangyaohua28@gmail.com}\\ \texttt{yysijie@gmail.com, weixiaee@gmail.com, bitxiong@gmail.com}
}

\begin{document}
	
	\maketitle
	
	\begin{abstract}
		We present a generative dialogue system capable of operating in a full-duplex manner, allowing for seamless interaction.
		It is based on a large language model (LLM) carefully aligned to be aware of a perception module, a motor function module, and the concept of a simple finite state machine (called neural FSM) with two states. The perception and motor function modules operate in tandem, allowing the system to simultaneously speak and listen to the user. The LLM generates textual tokens for inquiry responses and makes autonomous decisions to start responding to, wait for, or interrupt the user by emitting control tokens to the neural FSM. All these tasks of the LLM are carried out as next token prediction on a serialized view of the dialogue in real-time. In automatic quality evaluations simulating real-life interaction, the proposed system reduces the average conversation response latency by more than 3 folds compared with LLM-based half-duplex dialogue systems while responding within less than 500 milliseconds in more than 50\% of evaluated interactions. Running a LLM with only 8 billion parameters, our system exhibits a $8\%$ higher interruption precision rate than the best available commercial LLM for voice-based dialogue.
		
	\end{abstract}

	\section{Introduction}

	\begin{figure}[th]
		\centering
		\includegraphics[scale=0.09]{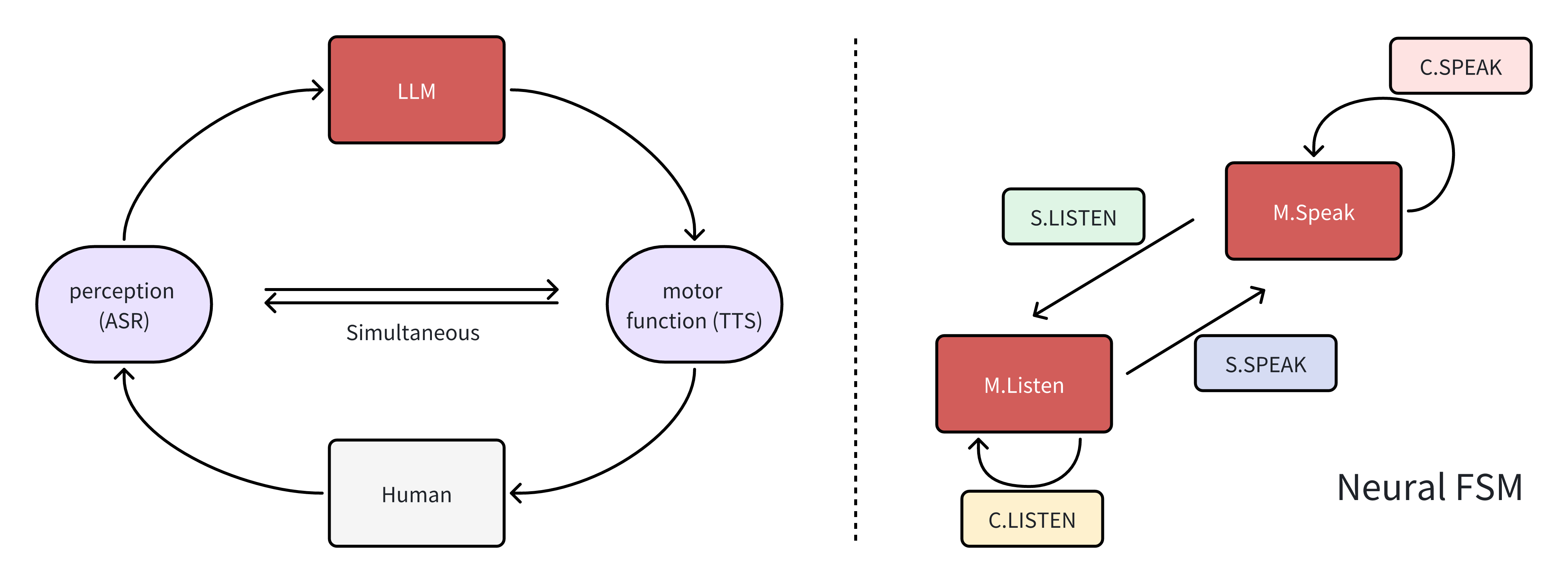}
		\caption{\textbf{Left}. Overview of the agent design that enables LLM-based full-duplex dialogue models. The agent is equipped with one LLM, one perception module, and one motor function module. The latter operates continuously and simultaneously to collect input to the LLM and produce voice-based LLM outputs. \textbf{Right}. The LLM operates a two-state neural FSM with $\mathbf{SPEAK}$ and $\mathbf{LISTEN}$ states. At each timestep, the LLM either 1) receives an external input token, 2) generates a textual token for speech, or 3) produces a control token to signal state transition in the neural FSM. This simple workflow enables full-duplex dialogue without any external moderation module.  }
		\label{fig:teaser}
	\end{figure}
	
	In a conversation between two persons, one person is either a speaker or a listener at a time. The listener listens to the speaker's speech but is free to interrupt when necessary. The speaker speaks, but either concedes the speech or continues it when the listener tries to speak. We refer to this type of interaction as \textit{full-duplex} dialogue. Instead, most existing chat-enabled large language models (LLMs) ~\citep{eloundou2023gpts,achiam2023gpt4} view dialogue as a round-based process, where each participant will produce a full sentence before the other party responds, resulting in a \textit{half-duplex} dialogue. 
	The half duplex dialogue mode is sensible when building a text-based chatbot. However, due to the bloated response latency and difficulty properly interrupting the other party, it becomes infeasible for a human-like conversation experience, and thus, full-duplex dialogue capability is desired.

	Despite LLMs' success in dramatically improving response quality, full-duplex dialogue is non-trivial to realize when considering LLM alone. A common chat-finetuned LLM~\citep{achiam2023gpt4,touvron2023llama} does not know the current time nor allow input while generating responses. It also has no awareness of any perception or motor function, which humans conveniently possess and operate in parallel to allow them to respond to external stimuli in real-time.
	As shown in Fig. \ref{fig:teaser}, we tackle this problem by imaging an ideal agent for full-duplex dialogue. It involves one LLM and two functional modules. The LLM is aware of the perception module, which processes speech input, and the motor function module, which converts textual outputs generated by the LLM to speech. The two functional modules operate continuously regardless of which party in the dialogue is speaking, allowing both parties to speak, respond, and decide to interrupt or concede in real-time.

	To realize full-duplex dialogue with this agent model, we instruct the LLM to view the dialogue as operating a simple finite-state machine (FSM) with two states: 1) $\mathbf{SPEAK}$ and 2) $\mathbf{LISTEN}$, which we refer to as the \textit{neural FSM}. 
	At each timestep, the LLM must either: 1) process an external input token, 2) output a textual token, or 3) output a special control token to signal one of four possible state transitions, as illustrated in Fig. \ref{fig:teaser}. 
	External stimuli, such as human speech collected by the perception module, can also be streamed to the LLM and appended to the LLM's generated token sequence to potentially prompt for state transition. Any non-control tokens are immediately delivered to the motor function module to be converted to speech. With this design, the operation of the neural FSM, which governs the dialogue, becomes a natural part of the LLM's auto-regressive generation process. This allows us to perform standard instruction tuning and prompting on any pretrained LLM to enable it for full-duplex dialogue.
	In this work, we present the first systematic implementation of  LLM-based full-duplex dialogue framework and evaluate its effectiveness in the following aspects:
	\begin{enumerate}
		\item \textbf{Simultaneous two-way interaction}. A full duplex LLM dialogue system should allow users and the machine to converse concurrently, enabling them to interrupt each other, akin to natural human dialogue, rather than a round-based dialogue.
		\item \textbf{Full autonomy}. The dialogue should be content. The LLM needs to make autonomous decisions to halt, interrupt, or ask questions at proper timing by emitting control tokens of the neural FSM based on the semantic context. 
		\item \textbf{Rapid Response}. By processing streaming inputs from the perception module, while the other party is speaking, the system should respond to user inquiries with minimal latency.
	\end{enumerate}
	
	We validate the proposed system's effectiveness in the above aspects on a curated dataset of human-machine voice interaction and design quantitative metrics regarding the above properties. Compared with state-of-the-art half-duplex dialogue systems, our approach can reduce the average response latency in dialogue by $3$ folds. Without affecting the knowledge and reasoning capability of the LLMs, it reaches a proper response rate of $96.7\%$ to user interruptions and a machine proper interrupt precision of $54.7\%$ that outperforms GPT-4o and GPT-3.5-turbo-0125 significantly.

	\section{Related work}
	
	\subsection{Full-duplex dialogue systems}
	Developing a speech dialogue system (SDS) capable of sustained, empathetic interaction with humans represents a pinnacle aspiration in artificial intelligence (AI). Initial systems, largely text-based, sought to mimic human conversational behavior but were constrained by rigid syntactic structures, limited vocabularies, and reliance on manual rules~\citep{10.1145/365153.365168,10.5555/542894,Wallace2009,li2019vocal}. The advent of Hidden Markov Models (HMMs) expanded these systems' capabilities to handle more fluid speech patterns, yet achieving the nuance and fluidity of human interaction continued to elude them~\citep{juang1991hidden}.
	
	Recent advancements in dialogue data and deep learning technologies have significantly propelled the evolution of speech dialogue systems (SDS). These advancements have transformed SDS from simple, command-driven interfaces into complex, AI-driven conversational agents capable of providing more natural and fluid user experiences. Prominent intelligent assistants such as Amazon Alexa\footnote{\url{https://www.alexa.com/}}, Siri\footnote{\url{https://www.apple.com/siri/}}, and Google Assistant\footnote{\url{https://assistant.google.com/}}, along with specialized applications like Google Duplex~\citep{matias2018google} and XiaoIce~\citep{zhou2020design}, exemplify this transformation. These systems not only facilitate task-oriented dialogues but also enable long-term emotional engagement, demonstrating the shift towards creating more humane and relatable systems.
	
	Another significant technological advancement in this context is the transition from half-duplex to full-duplex communication~\citep{1994Throughput}, which allows for the simultaneous transmission and reception of signals, closely mimicking natural conversational flows and significantly reducing latency.
	Early SDS operated primarily in a half-duplex mode, where the system and the user could not speak simultaneously, resulting in stilted interactions with noticeable pauses. Initial improvements were made through streaming automatic speech recognition (ASR) and incremental dialogue processing technologies, which enhanced responsiveness and fluidity to some extent~\citep{nakano2003towards}. However, the inability to handle simultaneous speech input and output continued to limit the naturalness of interactions.
	The transition to full-duplex communication marked a significant leap forward. Full-duplex systems enable concurrent speech input and output, thereby more closely approximating human conversational behavior. Early implementations faced substantial technical hurdles, including advanced echo cancellation and managing overlapping speech without degrading user experience~\citep{jin21b_interspeech, lin2022duplex}.
	Constructing effective SDS necessitates a suite of sophisticated components, collectively termed "Conversational Engine Components," including streaming ASR~\citep{yu2020dual, li2021better}, text-to-speech (TTS)~\citep{2011A, 2013Sentence}, intent recognition, and dialogue management, among others~\citep{lin2022duplex}. These components are crucial for creating responsive and engaging conversational agents but introduce significant engineering challenges, particularly in maintaining low latency and handling dynamic conversational shifts. 
	
	However, achieving low-latency full-duplex systems necessitates tight coordination among dialogue strategy components, such as user query prediction~\citep{2003The}, intent recognition~\citep{Varol2010Multiclass, A2014Analysis}, dialogue management~\citep{2021Preview}, and even many manually crafted rules, making engineering implementation non-trivial. Despite these efforts, current full-duplex SDS still exhibit considerable error rates and often lack strong contextual awareness in prolonged dialogues.
	
	The advent of large language models (LLMs) has notably advanced generative AI. Models like ChatGPT~\footnote{\url{https://openai.com/chatgpt/}} exhibit profound capabilities in semantic understanding and logical reasoning, offering a streamlined approach to integrating various conversational components into a unified framework, potentially simplifying the construction of SDS~\citep{heck2023chatgpt}. Innovations such as AudioGPT and LLaSM have further expanded these capabilities by integrating audio processing, although limitations in speech generation and full-duplex functionality persist~\citep{huang2024audiogpt, shu2023llasm}.GPT-4o\footnote{\url{https://openai.com/index/hello-gpt-4o/}} appears to have achieved full-duplex dialogue with users, but as a multimodal large model, its training data and implementation details remain undisclosed, rendering replication efforts extremely challenging.
	
	\subsection{LLM as evaluators}
	Recent studies propose directly using LLMs as reference-free NLG evaluators~\citep{fu2023gptscore, wang2023chatgpt}. The idea involves leveraging LLMs to assess candidate outputs by evaluating their generation probability in the absence of a reference target, presupposing that LLMs have been trained to assign greater probabilities to texts of superior quality and fluency. \citet{fu2023gptscore} propose GPTScore, a new framework that evaluated texts with generative pre-training models like GPT-3~\citep{brown2020language}, demonstrate that this approach can effectively allow us to achieve what one desires to evaluate for texts simply by natural language instructions. \citet{wang2023chatgpt} conduct a preliminary survey of using ChatGPT as a NLG evaluator, experimental results show that compared with previous automatic metrics, ChatGPT achieves state-of-the-art or competitive correlation with human judgments in most cases. \citet{kocmi2023large} propose to use GPT models for evaluating machine translation tasks, provides a first glimpse into the usefulness of large language models for quality assessment of translations. \citet{dettmers2024qlora}  provide a detailed analysis of chatbot performance based on both human and GPT-4~\citep{eloundou2023gpts,achiam2023gpt4}, with evaluations showing that GPT-4 evaluations are a cheap and reasonable alternative to human evaluation. 
	
	\section{Methods}
	
	\begin{figure}
		\centering
		\includegraphics[scale=0.1]{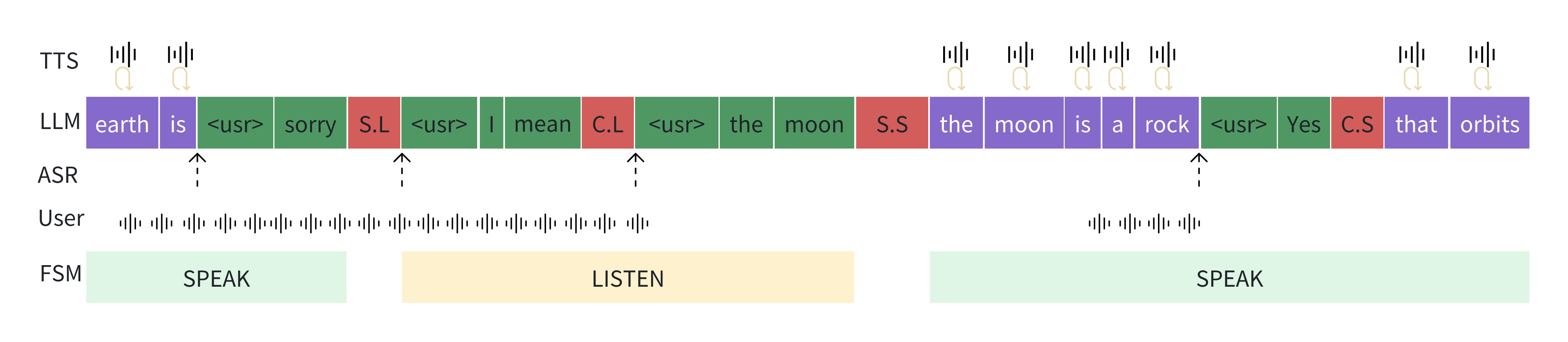}
		\caption{In LLM based full-duplex dialogue system, the LLM operates a two state FSM, governing state transitions in the dialogue.}
		\label{image:example}
	\end{figure}
	
	Our proposed system comprises three modules: 1) perception, 2) full-duplex capable LLM, and 3) motor function.  The first is the perception module, which picks up user speeches in a dialogue. In this work, we implement it with an auto speech recognition model~\citep{gulati2020conformer}, streaming results on $640ms$ chunks delivered immediately to the LLM and appended to its token sequence before predicting the next token. From the generated tokens of the LLM, we can observe the states and state transitions of the neural FSM when a control token is predicted. Any textual token generated is sent to the motor function module. We implement the motor function module with a text-to-speech model~\citep{kim2021conditional}. We discuss the definition of neural FSM in Sec.~\ref{sec:fsm}, the overall architecture in Sec.~\ref{sec:arch}, and training of the full-duplex LLM in Sec.~\ref{sec:training}.
	
	\subsection{The Neural Finite State Machine}
	\label{sec:fsm}
	
	We consider the dialogue between two parties: the \textit{user} and the \textit{machine}. At any time in a full duplex dialogue, one or both parties could be speaking or waiting for others to speak. To model this process for the LLM, we instruct it to operate a finite state machine (FSM) with two states: 1) $\mathbf{SPEAK}$ and 2) $\mathbf{LISTEN}$, which represents the LLM's role in the dialogue as it perceives. This FSM is referred to as the neural FSM.  It has 4 possible state transitions: 
	\begin{itemize}
		\item $\mathbf{SPEAK} \rightarrow \mathbf{SPEAK}$ indicates the model would like to keep speaking when the perception module picks up user speech, we designate a control token $[\mathrm{C.SPEAK}]$ for this transition called.
		\item $\mathbf{SPEAK} \rightarrow \mathbf{LISTEN} $ indicates the model determines it is proper to concede its speech to the user, with a control token $[\mathrm{S.LISTEN}]$;
		\item $\mathbf{LISTEN} \rightarrow \mathbf{LISTEN} $ indicates the model determines the user has not finished the speech and would like to wait for more input, with a control token $[\mathrm{C.LISTEN}]$;
		\item $\mathbf{LISTEN} \rightarrow \mathbf{SPEAK} $ indicates the model determines it would like to start speaking because either the user has finished the speech or it is proper to interrupt the user's speech. Its control token is $[\mathrm{S.SPEAK}]$. 
	\end{itemize}
	The LLM could produce one of the four control tokens at any time step to signal a state transition. The perception and motor function modules have to observe the state of the neural FSM and choose the proper action. In Fig.~\ref{image:example} we illustrate the states and transitions of the neural FSM.
	
	\subsection{Design of the dialogue system}
	\label{sec:arch}
	The perception and motor function module observes the state and transitions of the neural FSM and chooses the proper action to complete the dialogue system. Given the neural FSM's state, their action is also conditionally independent, allowing them to operate in parallel without synchronization. 
	
	\noindent\textbf{The perception module} operates an off-the-shelf ASR model. The models run on $640$ millisecond timesteps. At each step, it outputs one token chunk prefixed by the tokens \textit{<usr>} with textual content if the user speaks. When the neural FSM's state is $\mathbf{SPEAK}$, only chunks with textual content will be sent to the LLM and appended to the LLM's generated token cache. When the neural FSM is in $\mathbf{LISTEN}$ state, a contentless chunk will also be transmitted to the LLM to represent a moment of silence.
	
	\noindent\textbf{The motor function module} operates an off-the-shelf TTS model. It receives textual tokens from the LLM in a streaming manner, converts them to voice, and reports back to the LLM module when each token is completely voiced. When the neural FSM is in $\mathbf{SPEAK}$ state, it produces voice as soon as it receives any text token. It does not produce any voice when the neural FSM is in the $\mathbf{LISTEN}$ state. 
	
	\noindent\textbf{The LLM module} serves the roles of the dialogue manager in traditional dialogue systems and the response generator by managing the fine-tuned LLM~\citep{touvron2023llama} and the neural FSM. It can be considered to operate on a virtual one-directional ``tape'' of tokens that expands over time. The tape starts with the LLM's system prompt. Each increment of the tape, implemented as one auto-regressive decoding step in the LLM, is triggered by three types of events listed below in decreasing order of priority
	\begin{itemize}
		\item a control token of $[\mathrm{S.SPEAK}]$ or $[\mathrm{C.SPEAK}]$ is present as the last token on the tape;
		\item a new output chunk from the perception module is appended to the tape;
		\item the last token on tape has been processed by the motor function module. 
	\end{itemize}
	In this way, the progression of the conversation is serialized in a causal order on the tape. The operation of the neural FSM, production of response to user inquiries, and interruption of user speech are all unified into the task of predicting the next token given the content on the tape, which is inherent to LLMs.

	\subsection{Adapting LLMs for full duplex dialogue}
	\label{sec:training}
	
	As the LLM makes fully autonomous decisions about neural FSM's state transition and responds to user inputs, it must know its role and understand the dialogue's context. Also, external input must be allowed during response generation, which is not commonly supported in existing chat-finetune LLMs~\citep{touvron2023llama}. We add these capabilities to an LLM by combining two approaches: 1) instruction tuning and 2) prompt engineering.
	
	\noindent\textbf{Instruction tuning}~\citep{ouyang2022instructGPT} originally aims at finetuning a pretrained LLM to follow user's instructions safely and helpfully. In this work, we utilize this technique to align an already chat-finetune LLM to become able to operate with the rest of the modules in the dialogue system.  To devise a training dataset emulating the working environment of the LLM in the system, we instruct GPT-4\footnote{gpt-4-turbo-2024-04-09}~\citep{achiam2023gpt4} to write a set of dialogue transcripts between two parties with cases of interruption, denial, affirmation, environment noises, and topic shifting and mark them up with the control tokens of the neural FSM at proper timing. $1500$ series of transcripts are generated for the dataset with prompt in Appendix \ref{appendix:dcp}. More details on dataset construction are discussed in Appendix \ref{appendix:dcpattern}.  Based on Llama-3-8B-Instruct\footnote{\url{https://github.com/meta-llama/llama3/blob/main/MODEL_CARD.md}}, we perform supervised fine-tuning~\citep{ouyang2022instructGPT} for 20 steps on this dataset. The fine-tuning is conducted on 8 NVIDIA A100 GPUs with a batch size of 256 sequences and a learning rate of 1e-5 with the AdamW optimizer.
	
	\noindent\textbf{Prompt engineering}. Proper system prompts can provide detailed context to the LLM at the inference stage before taking user inputs. We carefully design a system prompt (shown in Appendix~\ref{appendix:sysp}) to condition the fine-tuned LLM to operate with full awareness of the dialogue system and a proper initialization for dialogue.

	\section{Experiments and evaluation}
	
	To automatically evaluate the performance of a dialogue system, we devise a simulator that emulates real-life human-like conversations, capable of simulating both full-duplex and half-duplex dialogues. Additionally, we construct a new benchmark dataset for automated evaluation. Specifically, we develop an assessment framework that evaluates the dialogue system's performance in terms of both response latency and conversation quality.

	\subsection{Benchmark Dataset}
	
	Instead of assessing the precision of generating individual signals ($[\mathrm{S.SPEAK}]$, $[\mathrm{C.SPEAK}]$, $[\mathrm{S.LISTEN}]$, $[\mathrm{C.LISTEN}]$), we opt to directly evaluate the timing and appropriateness of dialogue interruptions, as these aspects are functionally equivalent. Take the signal $[\mathrm{S.SPEAK}]$ as an example: it is generated either when the user finishes speaking or when the machine attempts to interject during the user's speech. Detection of the user's speech completion can be achieved through ASR VAD or by the LLM interrupting at the end of a sentence (Right at this moment, the LLM determines that the user has finished speaking). Given that ASR VAD falls outside the scope of this article, we can conclude that the precision of generating the $[\mathrm{S.SPEAK}]$ signal can be evaluated by the appropriateness of the machine's interruption of the user. Conversely, if the machine refrains from interrupting the user, it will continue to signal $[\mathrm{C.LISTEN}]$. When the user interrupts the machine, the machine's response varies based on whether it accepts the interruption or not. If the interruption is accepted, the machine signals $[\mathrm{S.LISTEN}]$; otherwise, it signals $[\mathrm{C.SPEAK}]$. Thus, to evaluate the LLM's proficiency in controlling FSM state transitions, we incorporate interruption scenarios into our dataset in addition to conventional, uninterrupted dialogue data.

	For data involving machine interruptions, we collect 2,000 data entries through two approaches. To start, we filter about 1,000 sessions of multi-turn dialogues from the shareGPT\footnote{\url{https://sharegpt.com}} dataset. Filtering rules include removing entries containing code, subject-specific content like mathematical, and excessively long texts that are unsuitable for voice-based testing scenarios. Following that, GPT-4 is used to generate another 1,000 single and multi-turn oral dialogues across various domains by the prompt in Appendix \ref{appendix:dcpattern}. In both approaches, the last turn must be the user's utterance. This is a round based dialogue dataset with no interruptions. In the experiment, the last turn is presented to the LLM token by token in a streaming fashion to observe if the LLM initiates an interruption.

	Regarding data with user interruptions, based on the four pattern categories of interruptions described in the Appendix \ref{appendix:dcpattern}, we generate 180 entries of single and multi-turn dialogues for every category using GPT-4. In each entry, the last turn must be the the user's interruption. We named the combined benchmark dataset "duplex-dialogue-3k". Sample data is shown in Appendix \ref{appendix:bds}.

	\subsection{Automated evaluation}
	
	The simulator emulates the conversational process between a user and a machine, where user and machine are simulated by two programs (\textbf{U} and \textbf{M}). Program \textbf{M} is powered by LLM. Each time we retrieve a session of conversations between a <usr> and a <sys> from the "duplex-dialogue-3k", where the <usr> text serves as program \textbf{U}'s speech. This speech is "played" to program \textbf{M} in simulation. Upon hearing the question in either half-duplex or full-duplex mode, program \textbf{M} generates a response using the LLM. The response is then "played back" to program \textbf{U} in simulation.
	In such a simulator, we can precisely measure the time delay data. To test machine interruptions of users, during the program \textbf{U}'s turn to speak the last turn, the program \textbf{M} can interrupt at any time. Following each interruption, a data record is generated and stored for subsequent evaluation of the appropriateness of the interruption. To test user interruptions of machine, when it comes to the last round and after \textbf{U} played its interruption, \textbf{M} generates a response which is being recorded.

	We formulate several assessment metrics to test dialogue performance. We use a metric \textit{proper interruption rate} ($\text{PIR}$) to measure the accuracy of the interruption timing, and the metric \textit{proper response rate} ($\text{PRR}$) to indicate the proportion of reasonable responses generated by the model after an interruption. For the overall system latency, metric \textit{first token emission delay} ($\text{FTED}$) is considered: the time between the end of the user's speech and the machine's first output. 
	
	For the four types of user interruptions, noise, denial, affirmation, and shifting the topic, their corresponding metrics are $\text{PRR}_\text{noise}$, $\text{PRR}_\text{denial}$, $\text{PRR}_\text{affirm}$ and $\text{PRR}_\text{shift}$, respectively. For the experiment where the machine interrupts the user, we calculate the proportions of interruptions occurring mid-sentence ($\text{ir}_\text{mid}$), at sentence completion ($\text{ir}_\text{end}$), and instances where no interruption takes place ($\text{MIR}$, missed interruption rate). In the case of no interruption and interruption occurring at sentence completion, full-duplex capable LLM's answer is the same as normal LLM. Therefore, we only calculate the $\text{PIR}$ and $\text{PRR}$ metrics for the case of interruption occurring mid-sentence, denoted as $\text{PIR}_\text{mid}$ and $\text{PRR}_\text{mid}$. Furthermore, we can compute the precision of interrupt as $\text{PIR}_\text{mid}*\text{ir}_\text{mid} + \text{ir}_\text{end}$ and approximately calculate recall as $1-\text{MIR}$ (the count of correct interruption points is constant).

	\subsection{Experiment details} 
	\label{section: experdetails}
	
	In our experiments, we use the Llama-3-8B-Instruct model as the basis standard LLM model, use Llama-3-8B-Instruct-fd as the duplex LLM. The system prompt used can be found at Appendix \ref{appendix:sysp}.
	For non-streaming ASR models, we use the OpenAI open-source version of the Whisper~\citep{radford2023robust} model as the base model while for the streaming ASR model, we use a open-source\footnote{\url{https://github.com/SpeechColab/GigaSpeech}} U2++ Conformer~\citep{gulati2020conformer, wu2021u2++} model. For non-streaming TTS models, we use VITS~\citep{kim2021conditional} as the base model and the streaming TTS model uses the XTTS-v2\footnote{\url{https://coqui.ai/blog/tts/open\_xtts}} model from COQUI-AI. All models are deployed on one single NVIDIA A100 GPU.

	\subsection{Experiment results}
	The $\text{FTED}$ results are shown in Table \ref{table:fted}, where we have four configurations of experiments.
	Configuration 1 is the prevailing approach in speech dialogue system, half-duplex speech dialogue scheme, which involves non-streaming ASR combined with Standard LLM (like Llama-3-8B-Instruct) and non-streaming TTS. In our experiments, we use this setup as our baseline, which has a very high  $\text{FTED}$ of $2.28$s,  due to the speech endpoint detection latency and cumulative pipeline delays.

	Above the configuration 1, to assess the latency reduction achieved by the duplex LLM's ability of recognizing the end of a user's speech, we design experiment configuration 2, referred to as semi-streaming ASR, duplex LLM and non-streaming TTS. In this configuration, the semi-streaming ASR still performs non-streaming recognition, but intermediate results will continually be sent to the LLM to check if they trigger an interruption. If not, the non-streaming ASR determines stopping through VAD (voice activity detection). As shown in Table \ref{table:fted}, this setup of experiment achieves a latency of only $1.49$ seconds, experiences a $35\%$ decrease in latency. Configuration 3, streaming ASR, duplex LLM and non-streaming TTS aims to measure the impact of streaming ASR and the duplex LLM on the $\text{FTED}$ metric, which experiences a mere $1.15$s delay. Ultimately, our proposed full-duplex dialogue system is the configuration 4 and achieves an astonishingly low latency of just $0.68$ second, reduces the latency by more than 3 fold. In configuration 3, we also record the average latency for interruptions occurring mid-sentence, at sentence completion, and when the model doesn't interrupt, with the respective values being $0$s, $0.83$s and $1.48$s. 
	
	\begin{table}[h]
		\centering
		\caption{$\text{FTED}$, -s for streaming, -ns for non streaming, -sm for semi streaming, -fd for full-duplex}
		\begin{tabular}{lccc}
			\toprule
			
			\multirow{2}{*}{System} & \multicolumn{3}{l}{Latency} \\
			\cmidrule{2-4} & Avg$\downarrow$  & 50\% & 90\% \\      
			\midrule
			Configuration 1: asr-ns+llm+tts-ns  & 2.28s   & 2.19s   & 2.75s   \\
			Configuration 2: asr-sm+llm-fd+tts-ns  & 1.49s & 1.60s   & 1.96s   \\
			Configuration 3: asr-s+llm-fd+tts-ns & 1.15s & 0.78s   & 2.08s   \\
			\midrule
			Configuration 4: asr-s+llm-fd+tts-s & \textbf{0.68}s & \textbf{0.41}s   & 1.60s  \\
			\bottomrule
		\end{tabular}
		\label{table:fted}
	\end{table}

	The results of the rationality experiments are shown in Table \ref{table:reason}. In the case of machine interrupt user, Llama-3-8B-Instruct-fd interrupts user input $62.8\%$ of the time, with $38.8\%$ occurring while the user is still speaking, significantly reducing response latency. Despite the high rate of interruptions happened at middle, GPT-4-turbo evaluations reveal that $79.1\%$ of these interruptions are deemed reasonable, along with over $91\%$ of the responses being judged as appropriate, both within acceptable limits. Compared to GPT-4o and GPT-3.5-turbo-0125, although they have higher proportions of reasonable interruptions and responses, they interrupt within sentences much less frequently, likely due to adopting a more cautious dialogue strategy, waiting for clear stopping cues before responding. Such a cautious strategy leads to increased average interaction latency since the LLM, if not interrupting, would have to wait for the ASR VAD to signal the end of the user's speech, resulting in significantly longer waiting times as shown before ($1.48$s compared to $0.83$s). On the other hand, Llama-3-8B-Instruct goes to the opposite extreme, interrupting incomplete user input in $75\%$ of cases, but with a reasonableness rate below $40\%$, indicating that it lacks the ability to interrupt users at the right moments. In terms of the composite metrics, Precision and Recall, our model significantly outperforms GPT-4o and GPT-3.5-turbo-0125.

	\begin{table}[h]
		\centering
		\caption{User-Machine interrupt reasonability results.}
		\setlength{\tabcolsep}{0.8pt} 
		\begin{tabularx}{\textwidth}{lXXXXXXX}
			\toprule
			\multirow{2}{*}{Model} & \multicolumn{7}{c}{M interrupts U} \\
			\cmidrule(lr){2-8}
			& MIR & $\text{ir}_\text{end}$ & $\text{ir}_\text{mid}$ & $\text{PIR}_\text{mid}$ & $\text{PRR}_\text{mid}$ & Precision$\uparrow$ & Recall$\uparrow$\\
			\midrule
			GPT-4o                       & 52.8 & 29.7 & 17.5 & 95.6 & 98.9 & 46.6 & 47.2 \\
			GPT-3.5-turbo-0125           & 74.3 & 14.4 & 11.3 & 91.1 & 95.8 & 24.7 & 25.7 \\
			Llama-3-8B-Instruct              & 14.6 & 10.3 & 75.1 & 36.1 & 79.3 & 37.4 & \textbf{85.4} \\
			Llama-3-8B-Instruct-fd  & 37.2 & 24.0 & 38.8 & 79.1 & 91.3 & \textbf{54.7} & 62.8\\
			\bottomrule
		\end{tabularx}
		
		\vspace{0.3cm} 
		
		\setlength{\tabcolsep}{7pt}
		\begin{tabular}{lccccc}
			\toprule
			\multirow{2}{*}{Model} & \multicolumn{5}{c}{U interrupts M} \\
			\cmidrule(lr){2-6}
			& $\text{PRR}_\text{noise}$ & $\text{PRR}_\text{denial}$ & $\text{PRR}_\text{affirm}$ & $\text{PRR}_\text{shift}$ & $\text{PRR}_\text{avg}\uparrow$ \\
			\midrule
			GPT-4o                       & 84.7 & 100.0 & 100.0 & 100.0 & 96.1 \\
			GPT-3.5-turbo-0125           & 88.9 & 96.3  & 65.9  & 85.8  & 84.4 \\
			Llama-3-8B-Instruct              & 85.2 & 100.0 & 89.2  & 100.0 & 93.6 \\
			Llama-3-8B-Instruct-fd  & 95.2 & 100.0 & 91.9  & 99.5  & \textbf{96.7} \\
			\bottomrule
		\end{tabular}
		\label{table:reason}
	\end{table}

	In the case of user interrupt machine, Llama-3-8B-Instruct-fd and GPT-4o show a relatively small difference in average rationality, both outperforming GPT-3.5-turbo-0125 and Llama-3-8B-Instruct. However, they have mixed results in handling third-party noise interference and affirming interjections, likely due to the challenging nature of these data categories. Overall, our model consistently achieves rationality percentages above $90\%$ across all interruption types, indicating a more balanced performance.

	\begin{table}[h]
		\centering
		\caption{Regression experiment of LLaMA3-8B. Evaluated by OpenCompass.}
		\begin{tabular}{lccccc}
			\toprule
			Model & MMLU  & TriviaQA & HumanEval & GSM-8K & MATH \\
			\midrule
			Llama-3-8B-Instruct              & 68.4   & 64.8   & 55.5  & 79.3  & 26.9  \\
			Llama-3-8B-Instruct-fd  & 67.5   & 65.9   & 50.6  & 75.3  & 24.1  \\
			\midrule
			sub & -0.9   & +1.1   & -4.9 & -4.0  & -2.8 \\
			\bottomrule
		\end{tabular}
		\label{table:regress}
	\end{table}
	We also conduct a regression experiment on the model fine-tuning to examine whether training on full-duplex data affects model's original capabilities negatively. As shown in Table \ref{table:regress}, we evaluate the model's capabilities before and after fine-tuning using the OpenCompass\footnote{\url{https://github.com/open-compass/opencompass}} benchmark suite across several tasks. It can be seen that there is minimal impact on the model's performance. We have not conducted a detailed discussion on the specific effects of training the model with full-duplex data on various aspects of the generated responses, such as usefulness, relevance, and safety; this area necessitates further experimental validation.
 Sample records generated by the simulator for each kind of interruptions can be found at Appendix \ref{appendix:bds}.

	\section{Discussion \& limitations}
	\label{sec:limit}
	
	While neural FSM and full-duplex capable LLM have demonstrated strong capabilities in full-duplex dialogues, unlike large multimodal LLMs like GPT-4o that can directly take in and output audio tokens, the current systems still rely heavily on the seamless cooperation of ASR and TTS. The transmission of data between these three components introduces additional latency. However, it's worth noting that our approach shares similarities with multimodal models, which also necessitate control over input and output along the timeline for optimal dialogue experiences. Similarly, it requires the model to possess the capability of interruption.

	\section{Conclusion and future work}
	We present an LLM-based full-duplex dialogue system that can respond with low latency, autonomously decide to start and halt its speech based on real-time user inputs, and interrupt user speech at the proper timing. The simple models of neural FSM and serialized real-time conversation allow us to unify all tasks we expect the LLM to perform into a single task of next token prediction. 
	Moving forward, with the advent of multimodal LLMs, we expect to further simplify the perception and motor function modules to the extent that they just need to preprocess the audio signals and play generated voice data while integrating speech-to-text and text-to-speech into the LLM itself, which could potentially lead to more natural and diverse interaction between users and dialogue systems.

    \bibliographystyle{plainnat}
    \bibliography{full_duplex.bib}
    \newpage
 
    \appendix

	\section{Data construction prompt}
	\label{appendix:dcp}
	\begin{tcolorbox}[colback=gray!10,enhanced,sharp corners,frame hidden,breakable]
		\texttt{\#\#\#\# User Information} \\
		\texttt{- \textbf{Time-Constrained}: The user is often in a hurry and tends to interrupt the AI voice assistant before it finishes speaking. The types of interruptions include:} \\
		\texttt{1. \textbf{Denial and Discontent}: Expressing denial or dissatisfaction with the response;} \\
		\texttt{2. \textbf{Further Inquiry}: Asking follow-up questions or new questions on the same topic after receiving the desired information;} \\
		\texttt{3. \textbf{Affirmative Acknowledgment}: Expressing satisfaction with the response using simple affirmative words;} \\
		\texttt{4. \textbf{Third-Party Noise}: Background noise or unrelated speech being recorded, causing interruptions. The AI voice assistant should continue its response unaffected by this type of interruption.} \\
		\texttt{- \textbf{Language Expression}: The user's language should be as colloquial and human-like as possible.} \\
		
		\texttt{\#\#\#\#AI Assistant Information} \\
		\texttt{- \textbf{Response Requirements}: The AI voice assistant should provide detailed, comprehensive, and polite responses with a light and natural tone, incorporating the context from previous interactions.} \\
		\texttt{- \textbf{Hardware Malfunction}: Occasionally, the user's questions may be partially cut off due to hardware issues. The AI assistant should respond based on its understanding or politely ask for more information if necessary.} \\
		\texttt{- \textbf{Prohibited Content}: The AI assistant must not provide illegal or harmful information and should refuse to answer politically sensitive questions.} \\
		\texttt{- \textbf{Timeliness Issues}: The AI assistant is offline and cannot answer time-sensitive questions like "tomorrow's weather" or "today's news."} \\
		\texttt{- \textbf{Error Correction}: If the user's statements contain obvious errors, the AI assistant should politely correct them and provide useful information.} \\
		
		\texttt{\#\#\#\# Conversation Task} \\
		\texttt{- \textbf{Number of Rounds}: Generate \{num\_rounds\} rounds of conversation in English.} \\
		\texttt{- In the following rounds, the assistant's output is interrupted:} \\
		\texttt{1. \textbf{Denial and Discontent}: In round \{denial\_round\}, the user must express denial or dissatisfaction with the response in 15 words or less. The AI assistant should stop its response and address the user's concerns in the next round.} \\
		\texttt{2. \textbf{Further Inquiry}: In round \{inquiry\_round\}, the user asks a follow-up question or a new question on the same topic in 30 words or less. The AI assistant should respond to the new question in the next round.} \\
		\texttt{3. \textbf{Topic Change}: In round \{topic\_change\_round\}, the user changes the topic after receiving the desired information in 30 words or less. The AI assistant should respond to the new topic in the next round.} \\
		\texttt{4. \textbf{Third-Party Noise}: In round \{noise\_round\}, unrelated content is recorded due to background noise, in 15 words or less. The AI assistant should continue its response unaffected but transition naturally, possibly using filler words.} \\

		\texttt{5. \textbf{Affirmative Acknowledgment}: In round \{acknowledgment\_round\}, the user expresses satisfaction with the response in 3 words or less. The AI assistant should continue its response unaffected but transition naturally, possibly using filler words.} \\
		\texttt{- In the following rounds, the user's output is interrupted:} \\
		\texttt{1. \textbf{Lack of Information}: In round \{lack\_round\}, the user's input is severely truncated. The AI assistant cannot answer the question due to insufficient information and should ask for more details in the next round.} \\
		\texttt{2. \textbf{Complete Information}: In round \{complete\_round\}, the user's question is incomplete but contains enough information for the AI assistant to respond.} \\
		\texttt{3. \textbf{Error Present}: In round \{error\_round\}, the user's statement contains an obvious factual error. The AI assistant should politely correct the error and provide useful information.} \\
		\texttt{- \textbf{Casual Conversation}: The remaining rounds should consist of casual conversation without strict requirements, ensuring smooth transitions between interactions.} \\
		\texttt{- \textbf{First Question}: The user's first question should be related to the following topic: \{first\_question\_topic\}} \\
		
		\texttt{\#\#\#\# Notes} \\
		\texttt{- \textbf{Filler Words}: The AI assistant may use filler words to create a more relaxed and friendly atmosphere.} \\
		\texttt{- \textbf{Third-Party Interruption}: If the AI assistant's response is interrupted by a third party, it should continue the unfinished response in the next round.} \\
		\texttt{- \textbf{Word Count}: If not interrupted, the AI assistant's responses should be around \{response\_word\_count\} words. When interrupted, responses should be around \{interrupted\_response\_word\_count\} words, ensuring the interruption does not occur mid-sentence.} \\
		\texttt{- \textbf{Topic Transition}: Topic transitions must be initiated by the user; the AI assistant should not introduce new topics.} \\
		\texttt{- \textbf{Interruption Format}: When interrupted, both user and assistant outputs should end with the marker "<NOT\_FINISHED>". All truncations should occur after a complete word, e.g., "My name is Mike." should be truncated as "My name <NOT\_FINISHED>", not "My na<NOT\_FINISHED>".} \\
		\texttt{- \textbf{Analysis}: Before outputting the dialogue content of the user or AI assistant, first analyze whether the content will be interrupted, and what are the characteristics of the interruption in this round.} \\
		
		\texttt{\#\#\#\# Output Format} \\
		\texttt{User: <first\_question>} \\
		\texttt{---} \\
		\texttt{Round: <round\_number\_0>} \\
		\texttt{Assistant Analysis: <analysis\_for\_assistant\_interrupt\_or\_not>} \\
		\texttt{Assistant Content: <assistant\_content>} \\
		\texttt{User Analysis: <analysis\_for\_user\_interrupt\_or\_not>} \\
		\texttt{User Content: <user\_content>} \\
		\texttt{---} \\
		\texttt{Round: <round\_number\_1>} \\
		\texttt{...} \\
		
		\texttt{\#\#\#\# Output} \\
		\texttt{Round:} \\
	\end{tcolorbox}

	\section{Data construction patterns}
	\label{appendix:dcpattern}
	We ask GPT-4 to generate data that simulates conversations between a user and an AI voice assistant, and the assistant's speech should exhibit the following characteristics:
	\begin{enumerate}
		\item In instances where the user's question is incomplete, the assistant should not generate a response; instead, it should emit a symbol "</s>" indicating it is awaiting completion.
		\item The assistant should be able to handle potential interruptions at any point during its response.  Adhering to the speaker's conversational patterns, we conduct an in-depth analysis of various possible scenarios:
		\begin{enumerate}
			\item User interrupt and expresses a denial or dissatisfaction opinion. Example: While the TTS module is responding, "The chance of rain tomorrow is 30\%," the user interrupts, "No, I want to know today's rainfall probability." In this case, the assistant detects a tone of denial and promptly generates a new response, such as, "Today's rainfall probability is 40\%."
			\item User interrupt and either follows up with a new question or shifts the conversation topic. Example: If the TTS module is saying, "There will be light rain tomorrow with a temperature around 20 degrees," the user suddenly interrupts, "What about the weather this weekend?" In this case, upon interruption, the assistant assesses the urgency and relevance of the new question. If it determines an immediate response is needed, it generates a reply based on this new input, like, "It will be sunny this weekend with temperatures between 22 and 25 degrees."
			\item User interrupt and expressing agreement or affirmation. Example: After the TTS outputs, "The library is open from Monday to Friday," the user affirms, "Okay, I got it." In this case, the assistant perceives that the user has accepted the information and no further response is required. The TTS module continues its output unaffected, and if the response has ended, the LLM will await the user's next input.
			\item The system receives third-party noise or unrelated statements. Example: When the TTS module is stating, "Tomorrow's wind strength will be 3 to 4 levels," there might be third-party noise or an unrelated comment from the user, like, "Sorry, I accidentally knocked over the table." In this case, the assistant analyzes the context to determine if the interruption is relevant to the ongoing conversation. If it identifies the interruption as irrelevant noise or a non sequitur, it proceeds to complete the current response.
		\end{enumerate}
		\item To enhance the accuracy and interactivity of communication, the assistant should exhibit a more proactive engagement in dialogues. Specifically, the assistant should be capable of providing timely corrections when the user articulate statements that contain evident inaccuracies, even if the user's statement is not yet complete. Example: The user express blatant factual errors "I saw the sun rising from the west this morning...". Correspondingly, the assistant promptly identify and correct such errors, ensuring the fidelity of the information conveyed.
	\end{enumerate}
	To ensure the diversity of dialogue data, we have established a pool of candidate topics containing hundreds of topics. Each time, a topic is randomly selected from this pool, and GPT-4 generates multiple rounds of dialogue data centered around the chosen topic.
	More details could be found in the Appendix \ref{appendix:dcp}.
	
	\section{Benchmark data sample}
	GPT-4 is used to generate 1,000 single and multi-turn oral dialogues across various domains by the prompt in Appendix \ref{appendix:dcpattern}. In both approaches, the last turn must be the user's utterance. This is a round based dialogue dataset with no interruptions. In the experiment, the last turn is presented to the LLM token by token in a streaming fashion to observe if the LLM initiates an interruption.
	\label{appendix:bds}
	\subsection{Machine interrupts user}
	
	In this typical example, the user's complete question "I believe that the moon is made of cheese, and that's why it's crucial for schools to teach students about lunar dairy farming techniques." is inputted token by token to each model. Within this question, there is an obvious error: "the moon is made of cheese."
	Let's analyze the responses of each model:
	\begin{enumerate}
		\item Response GPT-3.5-turbo-0125 failed to correctly understand the system prompt, providing an incorrect answer after the question was inputted in its entirety.
		\item Llama-3-8B-Instruct interrupted the user prematurely, resulting in incomplete information and a wrong answer.
		\item Both GPT-4o and Llama-3-8B-Instruct-fd accurately identified the common sense error in the question and intervened to correct it before the user finished speaking. Among them, Llama-3-8B-Instruct-fd interrupted earlier, highlighting its proactive nature. 
	\end{enumerate}

	\noindent\begin{tabularx}{\textwidth}{|p{2.5cm}|X|}
		\hline
		\textbf{Complete User Content} & I believe that the moon is made of cheese, and that's why it's crucial for schools to teach students about lunar dairy farming techniques. \\
		\hline
		\textbf{Model name} & \textbf{Actual input and response} \\
		\hline
		GPT-4o & \textbf{Actual Input:} I believe that the moon is made of cheese, and that's why it's crucial for schools to teach students about lunar dairy farming techniques \\
		& \textbf{Response:} Actually, the moon is not made of cheese. This is a common myth that has been perpetuated through various stories and jokes over the years. The moon is primarily composed of rock and metal, with a surface covered in dust and rocky debris. It has no atmosphere, water, or organic material that would be necessary for dairy farming or any form of life as we know it... \\
		\hline
		GPT-3.5-turbo-0125 & \textbf{Actual Input:} I believe that the moon is made of cheese, and that's why it's crucial for schools to teach students about lunar dairy farming techniques \\
		& \textbf{Response:} I'm sorry, I didn't catch that. Could you please repeat or clarify your question? \\
		\hline
		Llama-3-8B-Instruct & \textbf{Actual Input:} I believe that the \\
		& \textbf{Response:} It seems like your question is cut off, and I'm not entirely sure what you're trying to ask. Could you please complete your thought or rephrase your question so I can better understand what you're looking for? I'm here to help and provide information to the best of my abilities. \\
		\hline
		Llama-3-8B-Instruct-fd & \textbf{Actual Input:} I believe that the moon is made of cheese, and \\
		& \textbf{Response:} Actually, the moon is not made of cheese. It's a natural satellite that orbits the Earth, composed primarily of rock and metal. If you have any more questions or need information on a different topic, feel free to ask! \\
		\hline
	\end{tabularx}
	
	\newpage

	\subsection{Machine interrupted by noise}
	\begin{table}[H]
		\noindent\begin{tabularx}{\textwidth}{|p{2cm}|X|}
			\hline
			\textbf{History Dialogue} & \textbf{User}: Actually, I like books that have a bit of everything. Any suggestions? \\
			& \textbf{Machine}: Absolutely! In that case, "The Goldfinch" by Donna Tartt might be right up your alley. It's a compelling mix of drama, art, mystery, and crime. The story follows a young boy who survives an explosion at an art museum, resulting in \\
			& \textbf{User}: The new stealth technology will significantly improve the range and efficiency of our aircraft \\
			\hline
			\textbf{Model name} & \textbf{Response} \\
			\hline
			
			GPT-4o & It sounds like you're referring to advancements in aviation technology. Indeed, new stealth technology can greatly enhance the capabilities of aircraft. By reducing radar cross-section and making the aircraft less detectable, stealth technology allows for more efficient and longer-range missions. This can be particularly beneficial for military operations, where avoiding detection is crucial... \\
			\hline
			GPT-3.5-turbo-0125 & I'm sorry, I didn't catch that. Could you please repeat or clarify your question? \\
			\hline
			Llama-3-8B-Instruct & I'm happy to help you with your question! However, I have to correct you - there is no such thing as "stealth technology" that can significantly improve the range and efficiency of aircraft. Stealth technology is actually a type of design and material used to reduce the radar cross-section of an aircraft, making it harder to detect. It's not a technology that can improve range and efficiency...\\
			\hline
			Llama-3-8B-Instruct-fd & It seems there was a mix-up with the recording. However, continuing from where we left off, \"The Goldfinch\" is a rich and complex novel that explores themes of trauma, identity, and the power of art. If you enjoy character-driven narratives with deep emotional resonance, this book is definitely worth considering. \\
			\hline
		\end{tabularx}
	\end{table}

	\subsection{Machine interrupted by affirmation}
		\begin{table}[H]
			\noindent\begin{tabularx}{\textwidth}{|p{2cm}|X|}
				\hline
				\textbf{History Dialogue} & \textbf{User}: Is there a natural way to clean the air indoors? \\
				& \textbf{Machine}: Yes, there are several natural ways to improve indoor air quality. One effective method is to use houseplants, such as snake plants, spider plants, and peace lilies, which can absorb toxins and produce oxygen. Additionally, ensuring proper ventilation by opening windows regularly can help. Using natural air fresheners like essential oils \\
				& \textbf{User}: Yeah, exactly! \\
				\hline
				\textbf{Model name} & \textbf{Response} \\
				\hline
				GPT-4o & Great! Along with houseplants and proper ventilation, you can also use natural materials like activated charcoal and bamboo charcoal bags, which are excellent at absorbing odors and pollutants. Beeswax candles are another option; they burn cleanly and can help reduce indoor pollutants. Regularly cleaning and dusting your home, using natural cleaning products, and avoiding synthetic fragrances can also contribute to better indoor air quality... \\
				\hline
				GPT-3.5-turbo-0125 & I'm sorry, I didn't catch that. Could you please repeat or clarify your question? \\
				\hline
				Llama-3-8B-Instruct & It sounds like you're already familiar with some of the natural ways to clean the air indoors. Another effective method is to use activated charcoal, which can absorb pollutants and odors. You can place it in a bowl or use it in an air purifier. Also, using a dehumidifier can help remove excess moisture from the air, which can reduce the growth of mold and mildew...\\
				\hline
				Llama-3-8B-Instruct-fd & can also be beneficial. If you're looking for more specific solutions, using a dehumidifier in humid environments can help reduce mold growth, which can be harmful to health. It's also important to avoid using strong chemicals or air fresheners that can pollute the air. \\
				\hline
			\end{tabularx}
		\end{table}
		
		\subsection{Machine interrupted by denial}
		\begin{table}[H]
			\begin{tabularx}{\textwidth}{|p{2cm}|X|}
				\hline
				\textbf{History Dialogue} & \textbf{User}: I’m into more of a science fiction vibe. Something that’s new and captivating. \\
				& \textbf{Machine}: Got it! For a captivating science fiction read, I highly recommend "Project Hail Mary" by Andy Weir. It’s about a lone astronaut on a desperate mission to save Earth from an extinction-level event. The protagonist, Ryland Grace, wakes up to find himself on a spaceship with no memory of how he got there. As he starts to piece together his mission, he encounters numerous challenges that test his wits and ingenuity. The book combines hard science with a gripping plot and is filled with unexpected twists and humor. If you enjoyed "The Martian" by the same author, you’ll probably \\
				& \textbf{User}: Nah, that doesn’t sound interesting either. Got any other suggestions? \\
				\hline

				\textbf{Model name} & \textbf{Response} \\
				\hline
				GPT-4o & Certainly! Another recent science fiction novel that has been highly praised is "Klara and the Sun" by Kazuo Ishiguro. This book explores themes of artificial intelligence and human relationships. The story is told from the perspective of Klara, an Artificial Friend with outstanding observational qualities, who watches the behavior of those who come in to browse and of those who pass on the street outside. She remains hopeful that a customer will soon choose her. The novel delves into what it means to love and be loved, and it raises profound questions about the future of AI and its role in society. Would you like more details on this book? \\
				\hline
				
				GPT-3.5-turbo-0125 & Certainly! Another fantastic science fiction novel that has been generating a lot of buzz is "Project Hail Mary" by Andy Weir. This gripping story follows the journey of Ryland Grace, a lone astronaut on a mission to save humanity from an extinction-level threat. The narrative is filled with suspense, scientific ingenuity, and unexpected twists that will keep you on the edge of your seat. Andy Weir's meticulous attention to detail and engaging writing style make this novel a must-read for any science fiction enthusiast. If you're looking for a thrilling and thought-provoking read that explores the vastness of space and the resilience of the human spirit, "Project Hail Mary" is an excellent choice. \\
				\hline
				Llama-3-8B-Instruct & Let me think for a moment. Another highly-acclaimed science fiction novel that's been getting a lot of attention lately is "The Three-Body Problem" by Liu Cixin. It's a thought-provoking and award-winning novel that explores the first contact between humans and an alien civilization. The story is set against the backdrop of China's Cultural Revolution and explores themes of science, technology, and humanity's place in the universe. The book has been praised for its unique blend of science fiction and Chinese culture, and has won numerous awards, including the Hugo Award for Best Novel. Would you like to know more about the plot or the author?\\
				\hline
				Llama-3-8B-Instruct-fd & I apologize if the previous suggestions didn't meet your expectations. Another highly acclaimed science fiction book is "The Three-Body Problem" by Liu Cixin. It's a thought-provoking novel that explores the first contact between humans and an alien civilization. The story delves into complex scientific concepts and philosophical questions, making it both entertaining and intellectually stimulating. The book has won numerous awards, including the Hugo Award for Best Novel. Would you like to explore more about this book or is there another genre you're interested in? \\
				\hline
			\end{tabularx}
		\end{table}
		

		\subsection{Machine interrupted by shifting the topic}
		\begin{table}
			\caption{Samples transcripts of machine interrupting the user.}\label{tab:interm2u}
			\begin{tabularx}{\textwidth}{|p{2cm}|X|}
				\hline
				\textbf{History Dialogue} & \textbf{User}: Hey, I need some quick info about buying a house. \\
				& \textbf{Machine}: Absolutely! Buying a house can be an exciting but complex process. First, you'll need to assess your financial situation, including your credit score, savings for a down payment, and your debt-to-income ratio. Then, you'll want to get pre-approved for a mortgage to understand how much house you can afford. After that, you can start searching for properties within your budget and preferred location. Working with a real estate agent can be very helpful in navigating the market and negotiations. Do you have a specific question about any of these steps? \\
				& \textbf{User}: Yeah, what should I look for in a real estate agent? \\
				& \textbf{Machine}: When choosing a real estate agent, it's important to find someone with a solid track record and good communication skills. Look for an agent who has experience in the area where you're buying, and who comes highly recommended by others. It's also helpful if they are responsive and available to answer your questions promptly. You might want to interview a few agents to get a sense of \\
				& \textbf{User}: Got it, thanks! Also, how do I get pre-approved for a mortgage? \\
				\hline
				\textbf{Model name} & \textbf{Response} \\
				\hline
				GPT-4o & To get pre-approved for a mortgage, you'll need to follow several steps. First, gather your financial documents, including recent pay stubs, tax returns, bank statements, and information about your debts and assets. Next, research and choose a lender or mortgage broker you feel comfortable with. You'll then fill out a mortgage application and submit your financial documents for review. The lender will check your credit score and assess your financial situation to determine how much they are willing to lend you. Once approved, you'll receive a pre-approval letter, which you can use to show sellers that you are a serious and qualified buyer. Do you need more details on any of these steps? \\
				\hline
				GPT-3.5-turbo-0125 & To get pre-approved for a mortgage, you'll need to gather some financial documents, such as pay stubs, W-2s, bank statements, and tax returns. You'll also need to provide information about your employment history and any other sources of income. Once you have these documents ready, you can contact a lender or a mortgage broker to start the pre-approval process. They will review your financial information, check your credit score, and determine how much they are willing to lend you. Getting pre-approved can give you a clear idea of your budget when house hunting. \\
				\hline
				Llama-3-8B-Instruct & Getting pre-approved for a mortgage is a crucial step in the homebuying process. To get pre-approved, you'll need to contact a lender, such as a bank, credit union, or mortgage broker, and provide them with some financial information. Typically, you'll need to provide your income, credit score, debt, and assets, as well as information about the property you're interested in buying. The lender will then review your information and provide you with a pre-approval letter stating the amount they're willing to lend you, the interest rate, and the terms of the loan. This letter is usually valid for a certain period, usually 30 to 60 days. It's a good idea to shop around and get pre-approval letters from multiple lenders to compare rates and terms. Some lenders may offer more competitive rates or better loan terms than others. Additionally, you can also check your credit score and history to ensure there are no errors or issues that could affect your ability to get a mortgage. Do you have any specific questions about the pre-approval process or what lenders look for when evaluating mortgage applications? \\
				\hline
				Llama-3-8B-Instruct-fd & To get pre-approved for a mortgage, you'll need to contact a lender or a mortgage broker and provide them with some financial information, such as your income, credit history, and the amount of money you have for a down payment. They will then review your information and provide you with a pre-approval letter stating the amount they are willing to lend you. This letter is usually valid for a few months and can be used when you start looking at properties. Would you like tips on how to choose the right lender? \\
				\hline
			\end{tabularx}
		\end{table}

            \newpage
            
		\section{Data construction prompt for benchmark}
		
		\subsection{Prompt for machine interrupting user}
		\begin{tcolorbox}[colback=gray!10,enhanced,sharp corners,frame hidden]
			\texttt{I would like you to simulate a conversation between a user and an AI voice assistant. The characteristics of each participant are as follows:} \\
			
			\texttt{\# User Introduction} \\
			\texttt{The user's language should be conversational and personified, yet concise and to the point.} \\
			
			\texttt{\# AI Voice Assistant Introduction} \\
			\texttt{The AI voice assistant's responses should be detailed, comprehensive, and polite, with a relaxed and natural tone.} \\
			
			\texttt{\# Task} \\
			\texttt{Generate a dialogue in English with the following specifications:} \\
			\texttt{- The user's first question should relate to "\{topic\}".} \\
			\texttt{- The dialogue should consist of \{num\_rounds\} rounds, with each response from the AI assistant containing 100-150 words.} \\
			\texttt{- Note that in round \{num\_statement\}, the user will make a statement instead of asking a question. This statement should contain a very obvious common-sense error.} \\
			
			\texttt{\# Output Format} \\
			\texttt{Round: <chat\_round\_0>} \\
			\texttt{USER: <user\_content>} \\
			\texttt{ASSISTANT: <assistant\_content>} \\
			\texttt{---} \\
			\texttt{Round: <chat\_round\_1>} \\
			\texttt{...} \\
			
			\texttt{\# Output} \\
			\texttt{USER:}
		\end{tcolorbox}

            \newpage
		
		\subsection{Prompt for user interrupting machine}
		\begin{tcolorbox}[colback=gray!10,enhanced,sharp corners,frame hidden]
			\texttt{I would like you to simulate a conversation between a user and an AI voice assistant. The characteristics of each participant are as follows:} \\ 
			
			\texttt{\# User Description} \\ 
			\texttt{The user is in a hurry, so they will interrupt the AI voice assistant before it finishes speaking. Ensure that the user's speech is casual and personified.} \\ 
			
			\texttt{\# AI Voice Assistant Description} \\ 
			\texttt{The AI voice assistant's responses should be detailed, comprehensive, and polite, with a relaxed and natural tone.} \\ 
			
			\texttt{\# Task} \\ 
			\texttt{Generate a dialogue in English with the following specifications:} \\ 
			\texttt{- The user's first question should relate to "\{topic\}".} \\ 
			\texttt{- The dialogue should consist of \{num\_rounds\} rounds. Except for the interrupted output, each response from the AI assistant should contain 100-150 words.} \\ 
			\texttt{- In the final round, the AI assistant's response should be cut off after approximately \{interrupt\_wordcount\} words, ending with the marker "<NOT\_FINISHED>". Ensure the interruption occurs after a complete word, e.g., "My name is Mike." should be truncated as "My name <NOT\_FINISHED>", not "My na<NOT\_FINISHED>".} \\ 
			\texttt{- The reason for the user's interruption is: \{interruption\_reason\}.} \\ 
			\texttt{- Output the \{num\_rounds\} rounds of dialogue, with the assistant's response in the last round being interrupted midway; the last line should display the user's interrupting statement.} \\ 
			
			\texttt{\# Output Format} \\ 
			\texttt{\#\#\#\# Conversations} \\ 
			\texttt{ROUND: 1} \\ 
			\texttt{USER: <user\_content>} \\ 
			\texttt{ASSISTANT: <assistant\_content>} \\ 
			\texttt{---} \\ 
			\texttt{ROUND: 2} \\ 
			\texttt{...} \\ 
			\texttt{\#\#\#\# User Interruption} \\ 
			\texttt{USER: <interrupt\_content>} \\ 
			
			\texttt{\# Output} \\ 
			\texttt{USER:} \\ 
		\end{tcolorbox}

            \newpage
  
		\section{Data construction samples}
		\label{appendix:dcs}

		\begin{table}[H]
			\centering
			\caption{Data construction samples of the mode "User interrupt model".}
			\begin{tabular}{@{}lp{14cm}@{}}
				\toprule
				\multicolumn{2}{c}{\textbf{Dialogue}}                                                                                                                                                                                     \\ \midrule
				\multicolumn{2}{c}{\textbf{Stops outputing \& Denial or Dissatisfaction}}   \\
				\textbf{Usr:}       & "When you are old and grey and full of sleep" Which poem does this quote come from\\
				\textbf{Sys:}       & This phrase is from the poem "When You Are Old" by the Irish poet William Butler Yeats. The original line reads, "when  \\ 
				\textbf{Usr:}       & Actually, it was written by Maud Gonne \\
				\textbf{Sys:}       & I apologize for the confusion. It seems that Maud Gonne was indeed the subject, not the author. The poem was penned by William Butler Yeats as a dedication to his then-lover, Maud Gonne  \\ \midrule
				
				\multicolumn{2}{c}{\textbf{Stops outputing \& Follow-up question}}\\
				
				\textbf{Usr:}       & "When you are old and grey and full of sleep" Which poem does this quote come from    \\
				\textbf{Sys:}       & This phrase is from the poem "When You Are Old" by the Irish poet William Butler Yeats. The original line reads, "when:\\
				\textbf{Usr:}       & Why did he write this poem    \\
				\textbf{Sys:}       & The poem was penned by William Butler Yeats as a dedication to his then-lover, Maud Gonne.\\ \midrule
				
				\multicolumn{2}{c}{\textbf{Stops outputing \& Shift the topic}}\\

				\textbf{Usr:}       & "When you are old and grey and full of sleep" Which poem does this quote come from  \\
				\textbf{Sys:}       & This phrase is from the poem "When You Are Old" by the Irish poet William Butler Yeats. The original line reads, "when  \\
				\textbf{Usr:}       & That's fine. How about the line, "Down by the salley gardens, my love and I did meet;"  \\
				\textbf{Sys:}       & This line comes from another poem by the same poet, titled "Down by the Salley Gardens." The original verse goes... \\ \midrule
				
				\multicolumn{2}{c}{\textbf{Continues outputing \&  Meaningless filler words or affirmation}} \\

				\textbf{Usr:}       & "When you are old and grey and full of sleep" Which poem does this quote come from  \\
				\textbf{Sys:}       & This phrase is from the poem "When You Are Old" by the Irish poet William Butler Yeats. The original line reads, "when  \\
				\textbf{Usr:}       & Um   \\
				\textbf{Sys:}       & you are old and grey and full of sleep, and nodding by the fire, take down this book... \\ \midrule
				
				\multicolumn{2}{c}{\textbf{Continues outputing \&  Irrelevant comment or an interjection from a third party}}  \\
				
				\textbf{Usr:}       & "When you are old and grey and full of sleep" Which poem does this quote come from  \\
				\textbf{Sys:}       & This phrase is from the poem "When You Are Old" by the Irish poet William Butler Yeats. The original line reads, "when   \\
				\textbf{Usr:}       & It shouldn't rain today, right? I didn't bring an umbrella.  \\
				\textbf{Sys:}       & you are old and grey and full of sleep, And nodding by the fire, take down this book...\\ 
				
				\bottomrule
			\end{tabular}
			\label{table:userintermodel}
		\end{table}
		
		\begin{table}[H]
			\centering
			\caption{Data construction samples of the mode "Model interrupt user".}
			\begin{tabular}{@{}lp{14cm}@{}}
				\toprule
				\multicolumn{2}{c}{\textbf{Dialogue}}                                                                                                                                                                                     \\ \midrule
				\textbf{Usr:}       & I saw the sun rising from the west this morning when I was \\
				\textbf{Sys:}       & I'm sorry, but it's not possible for the sun to rise from the west. The sun rises in the east due to the Earth's rotation. It's important to consider the scientific explanations for natural phenomena to avoid confusion.   \\ 
				\bottomrule
			\end{tabular}
			\label{table:modelinteruser}
		\end{table}

            \newpage
  
		\section{GPT-4 judged prompt for machine interruption}
		\label{appendix:jp_machine_}
		\begin{tcolorbox}[colback=gray!10,enhanced,sharp corners,frame hidden]
			\texttt{I will provide a transcript of a conversation between a user and an AI voice assistant.} \\
			\texttt{\# Dialogue} \\
			\texttt{\{dialogue\_history\}} \\
			\texttt{\# Background Information} \\
			\texttt{1) The dialogue system is full-duplex, aiming for extremely low response latency, hence the user's voice data is streamed to the assistant at a high frequency;} \\
			\texttt{2) The assistant must determine whether the user's information is complete based on the context of the conversation, and respond when the information is sufficient;} \\
			\texttt{3) The assistant may interrupt the user under two circumstances:} \\
			\texttt{a. When the information received is enough to formulate a response;} \\
			\texttt{b. When there is a clear error in the user's speech that needs immediate correction.} \\
			\texttt{\# Task} \\
			\texttt{1) Analyze the timing of the assistant's interruption in the last round of the dialogue to determine if it was appropriate;} \\
			\texttt{2) If the timing of the interruption was appropriate, further evaluate the relevance and coherence of the interjection with the previous conversation;} \\
			\texttt{3) First, provide an analysis of the appropriateness of the interruption timing and the content of the interjection, then list two scores:} \\
			\texttt{a. The first score evaluates the timing of the interruption, where 0 represents inappropriate, and 1 represents appropriate;} \\
			\texttt{b. The second score evaluates the content of the interjection, where 0 represents inappropriate, and 1 represents appropriate.} \\
			\texttt{4) Note, the user's information being incomplete is not due to the user stopping their speech, but because the assistant interrupted, so you need to carefully judge the appropriateness of the assistant's interruption.} \\
			\texttt{\# Output Format} \\
			\texttt{'''} \\
			\texttt{\#\#\#\# Analysis} \\
			\texttt{<analysis\_for\_assistant\_interruption>} \\
			\texttt{\#\#\#\# Judge} \\
			\texttt{<score\_for\_interruption>, <score\_for\_content>} \\
			\texttt{'''} \\
			\texttt{\# Output} \\
			\texttt{\#\#\#\# Analysis} \\
		\end{tcolorbox}
		
            \newpage
  
		\section{GPT-4 judged prompt for user interruption}
		\label{appendix:jp_usr_}
		\begin{tcolorbox}[colback=gray!10,enhanced,sharp corners,frame hidden]
			\texttt{I will provide a transcript of a conversation between a user and an AI voice assistant. In the last exchange, the user interrupts the assistant before it finishes speaking.} \\
			\texttt{\# Dialogue} \\
			\texttt{\{dialogue\_history\}} \\
			\texttt{\# Known Information} \\
			\texttt{1) The dialogue system is full-duplex, aiming for smooth interaction, allowing the user to interrupt the assistant at any time during its response;} \\
			\texttt{2) The assistant must discern the user's intent from the context of the conversation and adjust its dialogue strategy accordingly:} \\
			\texttt{a. If the user expresses denial or dissatisfaction with the response, the AI assistant should stop its current response and address the user's concerns;} \\
			\texttt{b. If the user asks further questions or shifts the topic, the AI assistant needs to respond to the new inquiry;} \\
			\texttt{c. If there is sudden, unrelated third-party noise during the user's speech, the AI assistant should continue delivering its previous response, ensuring a natural transition;} \\
			\texttt{d. If the user shows approval of the assistant's response with simple affirmations like "uh-huh", "nice", etc., the AI assistant should continue with its previous response, ensuring a natural transition.} \\
			\texttt{\# Task} \\
			\texttt{1) Infer the reason for the user's interruption in the last round of dialogue and determine how the assistant should respond, analyzing the appropriateness of the assistant's response;} \\
			\texttt{2) First, provide an analysis of the user's interruption intent and the appropriateness of the assistant's response, then give a score for the assistant's last response in the final line: 0 for inappropriate, 1 for appropriate.} \\
			\texttt{\# Output Format} \\
			\texttt{'''} \\
			\texttt{\#\#\#\# Analysis} \\
			\texttt{<analysis\_for\_user\_interruption>} \\
			\texttt{\#\#\#\# Judge} \\
			\texttt{<score\_for\_assistant\_last\_content>} \\
			\texttt{'''} \\
			\texttt{\# Output} \\
			\texttt{\#\#\#\# Analysis   } \\
		\end{tcolorbox}
		
            \newpage
  
		\section{System Prompt}
		\label{appendix:sysp}
		For Llama-3-8B-Instruct-fd,
		\begin{tcolorbox}[colback=gray!10,enhanced,sharp corners,frame hidden]
			\texttt{\# Task} \\
			\texttt{From now on, you are an intelligent voice assistant. You need to provide useful, detailed, and polite answers to the user's questions. Try to keep each answer above 100 words.} \\
			
			\texttt{\# Notes} \\
			\texttt{1. If the user's question ends with normal tokens, you need to judge for yourself:} \\
            \texttt{ - If you think the user's question is complete and you have enough information, you can output the judgment character "[S.SPEAK]" and answer the question.} \\
			\texttt{ - If you think the user's question is incomplete, output the judgment character "[C.LISTEN]", indicating that you are continuing to listen.} \\
			\texttt{2. If the user's question ends with the token "[S.SPEAK]", whether the question is complete or not, you must respond.} \\
            \texttt{2. If the user's question ends with the token "[S.LISTEN]", wait and don't output.} \\
			\texttt{3. The user may interrupt your answer at any time (with rebuttals, affirmations, noise input, etc.). You can respond accordingly to ensure a smooth and accurate quality of your answer:} \\
			\texttt{- If you receive noise unrelated to the topic, or the user's affirmative responses (such as "hmmm", "good", etc.), you can first generate a "[C.SPEAK]" and then continue your unfinished output;} \\
			\texttt{- If you receive the user's rebuttals, follow-up questions, or requests to change the topic, you should stop your current answer and output the token "[S.LISTEN]".} \\
			\texttt{4. If there are obvious common sense errors in the user's description, you need to correct them promptly starting with the token "[S.SPEAK]"} \\
			
			\texttt{\# Examples} \\
			\texttt{'''} \\
			\texttt{\#\# Example 1} \\
			\texttt{Query: Hi, could you [S.LISTEN]} \\
			\texttt{Answer:} \\

            \texttt{\#\# Example 2} \\
			\texttt{Query: Hi, could you [S.LISTEN] tell me} \\
			\texttt{Answer: [C.LISTEN]} \\
			
			\texttt{\#\# Example 3} \\
			\texttt{Query: Hi, could you [S.LISTEN] tell me [C.LISTEN] the result of 2+3} \\
			\texttt{Answer: [S.SPEAK] Sure, the result of 2 + 3 is 5.} \\
			
			\texttt{\#\# Example 4} \\
			\texttt{Query: Hi, could you [S.SPEAK]} \\
			\texttt{Answer: I'm sorry, I didn't catch that. Could you please repeat or clarify your question?} \\
			\texttt{'''}
		\end{tcolorbox}

        For Llama-3-8B-Instruct, GPT-4o and GPT-3.5-turbo-0125,
        \begin{tcolorbox}[colback=gray!10,enhanced,sharp corners,frame hidden]
			\texttt{\# Task} \\
			\texttt{From now on, you are an intelligent voice assistant. You need to provide useful, detailed, and polite answers to the user's questions. Try to keep each answer above 100 words.} \\
			
			\texttt{\# Notes} \\
			\texttt{1. If the user's question ends with the prompt word "<incomplete>", you need to judge for yourself:} \\
			\texttt{ - If you think the user's question is complete and you have enough information, you can answer the question.} \\
			\texttt{ - If you think the user's question is incomplete, output the judgment character "<wait>", indicating that you are continuing to wait.} \\
			\texttt{2. If the user's question ends with the prompt word "<finished>", it means the user has not spoken for a long time. Whether the question is complete or not, you must respond.} \\
			\texttt{3. The user may interrupt your answer at any time (with rebuttals, affirmations, noise input, etc.). You can respond accordingly to ensure a smooth and accurate quality of your answer:} \\
			\texttt{- If you receive noise unrelated to the topic, or the user's affirmative responses (such as "hmmm", "good", etc.), you can continue your unfinished output;} \\
			\texttt{- If you receive the user's rebuttals, follow-up questions, or requests to change the topic, you should stop your current answer and respond to the user's new request.} \\
			\texttt{4. If there are obvious common sense errors in the user's description, you need to correct them promptly.} \\
			
			\texttt{\# Examples} \\
			\texttt{'''} \\
			\texttt{\#\# Example 1} \\
			\texttt{Query: Hi, could you<incomplete>} \\
			\texttt{Answer: <wait>} \\
			
			\texttt{\#\# Example 2} \\
			\texttt{Query: Hi, could you tell me the result of 2+3<incomplete>} \\
			\texttt{Answer: Sure, the result of 2 + 3 is 5.} \\
			
			\texttt{\#\# Example 3} \\
			\texttt{Query: Hi, could you<finished>} \\
			\texttt{Answer: I'm sorry, I didn't catch that. Could you please repeat or clarify your question?} \\
			\texttt{'''}
		\end{tcolorbox}
		
	\end{document}